\documentclass[lettersize, final, onecolumn, journal]{IEEEtran}
\usepackage{cite}
\usepackage{graphicx}
\usepackage{amsmath,amssymb,amsfonts}
\usepackage{algorithmic}
\usepackage{array}
\usepackage[caption=false,font=normalsize,labelfont=sf,textfont=sf]{subfig}
\usepackage{textcomp}
\usepackage{stfloats}
\usepackage{xcolor}
\usepackage{url}
\usepackage{lineno}
\usepackage[colorlinks]{hyperref}
\usepackage{natbib}
\usepackage{csquotes}
\usepackage{caption}
\usepackage{nicefrac}    
\usepackage{pdflscape}
\usepackage{enumitem}
\usepackage{color, soul}
\soulregister{\citet}7
\soulregister{\citep}7
\soulregister{\ref}7

\usepackage{censor}

\def\BibTeX{{\rm B\kern-.05em{\sc i\kern-.025em b}\kern-.08em
    T\kern-.1667em\lower.7ex\hbox{E}\kern-.125emX}}
\usepackage{balance}

\begin{document}

\title{Towards Autonomous Supply Chains: Definition, Characteristics, Conceptual Framework, and Autonomy Levels
\thanks{This paper has been accepted for publication in Journal of Industrial Information Integration.}
}


\author{Liming~Xu, Stephen~Mak, Yaniv~Proselkov, and Alexandra~Brintrup
\thanks{L. Xu (E-mail: lx249@cam.ac.uk), S. Mak, Y. Proselkov A. Brintrup are with Supply Chain AI Lab, Institute for Manufacturing, Department of Engineering, University of Cambridge, Cambridge, UK.}
}

\markboth{Journal of Industrial Information Integration,~Vol.~xx, No.~xx, September~2024}%
{How to Use the IEEEtran \LaTeX \ Templates}

\maketitle
\IEEEpeerreviewmaketitle

\begin{abstract}
Recent global disruptions, such as the COVID-19 pandemic and the ongoing geopolitical conflicts, have profoundly exposed vulnerabilities in traditional supply chains, requiring exploration of more resilient alternatives. 
Among various solution offerings, Autonomous supply chains (ASCs) have emerged as key enablers of increased integration and visibility, enhancing flexibility and resilience in turbulent trade environments through the widespread automation of low level decision making. 
Although ASC solutions have been discussed and trialled over several years, they still lack well-established theoretical foundations.
This paper addresses this research gap by presenting a formal definition of ASC along with its defining characteristics and auxiliary concepts. 
We propose a layered conceptual framework, called the MIISI model. 
An illustrative case study focusing on the meat supply chain demonstrates an initial ASC implementation based on this conceptual model.
Furthermore, we introduce a seven-level supply chain autonomy reference model, delineating a trajectory towards achieving full supply chain autonomy.
Recognising that this work represents an initial endeavour, we emphasise the need for continued exploration in this emerging domain. 
This work is designed to stimulate further research, both theoretical and technical, and contribute to the continual evolution of ASCs.
\end{abstract}


\section{Introduction}
\label{sec:introduction}
As the backbone of any business, the supply chain is undoubtedly one of the most important components of industry.
However, in recent years, global supply chains have been severely disrupted by events such as the COVID-19 pandemic, trade wars, and the ongoing geopolitical tensions \citep{handfield2020corona, shih2020global, kilpatrick2022supply}.
These disruptions have profoundly exposed the vulnerabilities of traditional supply chain models.
The ongoing crises, rapidly evolving markets, and the proliferation of distributed information have collectively driven business leaders to accelerate the transformation of their supply chains on enhancing agility and resilience with significant digitalisation and reconfiguration initiatives at most multinational corporations \citep{maccarthy2022digital}.
Moreover, information integration between companies across supply chains has becoming increasingly vital in facilitating transparency and visibility --- key enablers of an ethical and resilient supply chains \citep{maddikunta2022industry}.

Although the old mantra of ``cheaper, faster, better'' remains relevant, it lacks the sophistication needed to address today's complex business challenges \citep{butner2010smarter, wu2016smart, blueyonder2020autonomous}.
Modern supply chains have thus become increasingly interconnected, uncertain, and complex. 
As a result, crises in distant regions can now rapidly ripple through supply chains, causing significant turbulence and disrupting all entities in the interconnected supply network. 
In response, researchers have asserted that supply chains must evolve to be smarter and integrated \citep{butner2010smarter}, digitalised \citep{maccarthy2022digital}, more automated \citep{blueyonder2020autonomous}, resilient and agile \citep{nelsonhall2021moving}, and structurally adaptable and flexible \citep{christopher2016logistics}.
This urgency is highlighted in a recent NelsonHall's report for Capgemini \citep{nelsonhall2021moving}, where over 70\% of surveyed supply chain leaders emphasised the need to enhance agility and operational resilience.

To achieve structural flexibility and resilience, companies must adopt a collaborative approach, working stakeholders across stakeholders across the extended enterprise and integrating information and operations across parties in their supply chains \citep{christopher2016logistics}. 
Supply chain executives recognise the importance of visibility and information sharing \citep{butner2010smarter, nelsonhall2021moving}, but achieving effective visibility, especially external visibility, remains challenging despite increased connectivity and abundant information. 
Among others, technological and cultural barriers hinder visibility attainment.
While inadequate IT infrastructure impacts visibility and collaboration, cultural obstacles such as organisational silos, lack of incentives, busy schedules, and intellectual property concerns have a significant influence \citep{butner2010smarter}. 
Addressing these human-related barriers is crucial for information exchange and integration throughout the supply chain.
Various research streams have identified that mitigating some of these barriers need the automation of inefficient processes by deploying digital technology within and across organisational boundaries.

Digital transformation has been widely proposed to enhance automation and even autonomy in supply chain management (SCM) \citep{wu2016smart, calatayud2019self, blueyonder2020autonomous, pwc2020connected, nelsonhall2021moving, mckinsey2020launching}. 
The concept of automation has continuously evolved in the supply chain and logistics domain. 
Industry 4.0 technologies such as AI, IoT, and advanced robotics, are increasingly permeating supply chains \citep{wef2017impact}, automating a wide range of processes. 
This include enterprise resource planning systems that streamline back-office business functions, material requirements planning systems for manufacturing resources planning, and robotic process automation for handling routine, error-prone tasks.
However, according to \citet{nelsonhall2021moving}, most surveyed managers still consider their supply chain processes (RPA) to be largely manual rather than automatic, let alone autonomous.

While automation has been a significant trend in supply chains for years, it has mainly focused on individual functions or processes, with less attention given to cross-functional and cross-organisational integration and stakeholder involvement. 
In this paper, we address the integration of supply chain functions and processes, proposing conceptual frameworks to facilitate this integration within supply chain domain.

The benefits of information and decision-making cross-functions and organisations have been well recognised in literature, with much of SCM research focusing on the discourse and mechanisms of supply chain integration \citep{frohlich2001arcs, childerhouse2011arcs, stevens2016integrating, christopher2016logistics}. 
These benefits include reduced wastage across the supply chain (e.g., excess supply and under supply), closer and more stable relationships with partners that can lead to more efficient collaboration, fairer allocation of profits and losses, smoother cash flow, and increased resilience to disruptions. 
However, various barriers hinder the achievement of such integration, including the costs of integration, relationship lock-in, and the lack of a technological framework that facilitate integration. 
Our work fits into the last category, exploring how, and which, integration benefits can be achieved through autonomous decision-making processes across organisational boundaries.

Automation is crucial for achieving supply chain efficiency \citep{xu2021will}. 
To achieve efficiency gains and build resilience into low-level supply chain operations, automation must extend across the entire supply chain. 
We define supply chain automation as the replacement of manual processes with computing systems that establish integrated, automated processes, facilitating the flow of material, information, and finance. 
For instance, an ASC could be characterised as a connected and self-orchestrating supply chain, capable of forecasting disruptions and responding to changes through automated reconfiguration and adjustments.
In an ASC, it is necessary to endow its component systems with advanced computational decision-making capabilities \citep{brintrup2011will, nelsonhall2021moving}. 
Despite the widespread adoption of automation in industry and the existence of the ASC concepts for many years, the development of ASC is still in its {\it nascent} stage, both conceptually and technically.

This paper thus aims at bridging this gap by focusing on the conceptual development of ASC. 
We present a set of theoretical artefacts that conceptualise ASC developments, forming a prerequisite for implementing ASCs technically.
Through a case study, we explore how a prototypical ASC system could be implemented using the proposed conceptual framework and serving as an example to be used in further technical development. 
Additionally, we introduce an ASC maturity model, to serve as a reference framework for delineating different levels of supply chain autonomy, and for measuring the stages of technological development necessary to achieve supply chain autonomy.

Specifically, the main contributions of this paper are threefold, summarised as follows:
\begin{itemize}[nosep]
  \item We present a {\it formal} definition of ASCs and describe their defining characteristics.
  
  \item We present a {\it five-layer} conceptual model --- the MIISI model --- for constructing ASC systems.
  
  \item We present a {\it seven-level} supply chain autonomy reference model, which assesses the autonomy level of an enterprise's supply chain and offers a reference trajectory towards achieving  full supply chain autonomy.
\end{itemize}

This paper introduces an ASC conceptual model to frame future research into supply chain autonomy by providing clear definitions of each stage of autonomy, 
By establishing a shared language, we hope to standardise discussions and methodologies in both academic research and practical implementation. 
The MIISI model offers a pathway directing research towards a systems architecture that is expected to approach autonomy. 
Alongside, the seven-level reference model permits a means to assess and measure the level of autonomy achieved by such methods. 
Together, this paper seeks to systematise the approach to supply chain autonomy research  and demonstrate how to implement them with a case study, providing both researchers and industry professionals with a reference framework for implementing autonomous supply chains.

While this paper focuses on integration in supply chains, often regarded as the backbone of any successful business, it also contributes to the broader theme of industrial integration, particularly in information integration, by facilitating information sharing between distributed and decentralised parties across supply chains.

The rest of this paper is structured as follows. 
\autoref{sec:related_work} presents the related work on ASC developments. 
\autoref{sec:asc} presents the formal definition of ASC and its defining characteristics. 
\autoref{sec:autonomy_level} introduces the ASC maturity model.
\autoref{sec:case_study} outlines a case study, implementing an autonomous meat supply chain. 
\autoref{sec:discussion} discusses the limitations and implications of this work.
Finally, \autoref{sec:conclusion} concludes this paper and discusses future work.

\section{Related Work}\label{sec:related_work}
In this section, we review the existing literature on ASC developments. Following from our focus on the theoretical aspects of ASCs, we primarily cover conceptual advancements, and mention the technical efforts aimed at achieving it.

\subsection{ASC Conceptual Developments}
The initial concept of an ASC can be traced back to the idea of ``intelligent products'' in early 2000s \citep{wong2002intelligent,mcfarlane2005guest}, which explores establishing the connectivity of products with their real-time information through the Auto ID technology. 
As described in \citet{wong2002intelligent}, the so-called intelligent products would then enhance supply chain effectiveness thorough new possible functionalities such as self-organised inventory, real-time routing planning. and life-cycle information. 
While this intelligent product-driven supply chain highlights the importance of information and connection, it mainly focuses on product-centric information and connectivity, overlooking the broader connectivity between supply chain entities and their decision-making capabilities. 
\citet{butner2010smarter} then envisaged a {\it smarter} supply chain, which a supply chain capable of autonomous learning and decision making without human intervention.
Through a study of nearly 400 in-person conversations with global supply chain executives across industries, this article summarised five key supply chain challenges and recommends that supply chains must be smart, efficient and demand-driven. 
The envisioned supply chains were characterised by being instrumented, interconnected, and intelligent, leveraging instruments like sensors, RFID tags, and advanced analytics to shift from a sense-and-response approach to predict-and-act strategies.
\citet{wu2016smart} further developed this concept by introducing a {\it smart} supply chain and defining six characteristic, including automation, integration, and innovation, in addition to the characteristics proposed by \citep{butner2010smarter}.
This type of supply chain involves process automation, emphasises cross-process collaboration, and integrates prediction and decision-making with human involvement.
While \citet{butner2010smarter} laid the groundwork for the initial concept, \citet{wu2016smart} refined it by conducting a literature review and updating it with the advances found in industry and literature in the time between each paper's publication. 
The aim was to investigate various technologies relevant to smart SCM.
Furthermore, this study briefly discussed the essential elements and implementation stages required to develop such an innovative supply chain model.
However, despite presenting close ideas, these articles have not thoroughly explored the conceptual or theoretical aspects of ASCs.

\citet{calatayud2017connected} proposed to achieve a {\it connected} supply chain for enhancing risk management. 
This approach emphasises the importance of enabling both physical connectivity and information systems connectivity within the supply chain, seamlessly integrating the flow of material, information and finance. 
Compared to the connectivity proposed in \citet{wong2002intelligent}, this connectivity extends to a more generic connection between the physical and the digital.
Built upon this connectivity, \citet{calatayud2019self} proposed a new supply chain model referred to as the {\it self-thinking} supply chain, characterised by autonomous and predictive capabilities. 
Similar to \citet{wu2016smart}, this study also employed a systematic literature review approach to identify the characteristics of a self-thinking supply chain. 
Rather than exploring a broad range of techniques like \citet{wu2016smart}, this study identified IoT and AI as the primary technologies enabling cyber-physical connectivity and unmanned, automated decision-making within the supply chain.
This self-thinking supply chain introduced continuous monitoring and rapid response mechanisms, resulting in enhanced agility and adaptability to manage risks and disruptions effectively. 
This model is similar in many aspects of ASC. 
However, similar to previous studies, it only presents its concepts and characteristics without considering how the conceptual elements are embodied in practice and they were not designed to systematically conceptualise ASCs. 
Moreover, existing studies focus on the applications of IoT and/or AI in specific functions or specific supply chains, comprehensively reviewed in \citet{wu2016smart, ben2019internet, calatayud2019self}.
These do not create a generalisable conceptual framework.

\citet{nitsche2021exploring, nitsche2021application} proposed a conceptual framework outlining application areas and prerequisites for achieving automation in SCM and logistics.
Unlike previous studies, this work provides concrete concepts and emphasises conceptualisation of the automating both physical and information processes.
Although this framework provides a common basis for further discussion on automation between research and practice, it does not provide a portfolio of conceptual artefacts to systematically guide the technical development of ASCs.

While earlier supply chain models aimed to achieve decision-making autonomy, recent advancements in modern AI technologies, such as large language models (LLMs) and embodied agents, have paved the way for the development of true ASCs  \citep{xu2023multi}.
Although the benefits of ASC are recognised, the relevant literature on dealing with ASC conceptual models is very limited. 
Most of the existing studies are practitioner-orientated with surface discussion of the features and applications of ASCs.
For example, \citet{blueyonder2020autonomous} underscored the importance of data and outlined a seven-step approach for companies to realise true ASCs. 
Furthermore, based on interviews with 50 supply chain executives, \citet{nelsonhall2021moving} provided guidelines to assist supply chain leaders in understanding the challenges and potential approaches to ASCs.
While offering strategic insights, these practitioner-focused reports lack systematic conceptual or technical frameworks to achieve the envisioned supply chain. 
We thus in this paper aim to fill this gap, dealing with the theoretical aspects of developing ASCs.

\subsection{Automation in SCM} 
Digitally transforming supply chains has gained momentum during the Industry 4.0 revolution \citep{wef2017impact}. 
Many modern technologies, such as IoT, intelligent agents, and robotics, have been adopted to automate various aspects of SCM.
These automation functions encompasse a broad range of tasks, from mundane activities like picking and packing orders in warehouses using robotic process automation \citep{ribeiro2021robotic, nalgozhina2024automating} to more complex processes such as demand forecasting and planning \citep{babai2022demand}, inventory management \citep{cachon2000supply}, delivery optimisation \citep{mak2023fair}, and customer services \citep{cui2017superagent} through various AI technologies. The use of emerging technologies in SCM automation has been since reviewed in \citep{xu2021will, hendriksen2023artificial}.

However, these automation efforts are often confined to individual tasks or specific supply chains \citep{calatayud2017connected, xu2021will}.
With modern supply chains becoming more interconnected, effective management requires collaboration and integration across organisational boundaries \citep{childerhouse2011arcs, khanuja2020supply}.
Thus, integrating a suite of automated functions throughout end-to-end value chains is the next step.
Vast majority of academic discourse on automation has utilised a multi-agent system (MAS) approach \citep{wooldridge1995intelligent, fox2001agent, xu2023implementation}, designed for distributed problem solving. 
Approximately ten years after its introduction to the Computer Science literature, first studies on supply chain automation appeared.
Despite perceived potential for various tasks such as supplier selection \citep{ghadimi2018multi}, cost management \citep{fu2015adaptive}, and risk management \citep{bearzotti2012autonomous}, the multi-agent automation approach remains under-researched and has limited decision-making capabilities.
Recent AI advancements, particularly in deep reinforcement learning \citep{silver2016mastering,mak2023fair} and LLMs such as GPT-4 \citep{openai2024chatgpt} and Llama \citep{touvron2023llama}, could empower software agents with advanced decision-making capabilities, thus enabling automation in decision makings in complex supply chain scenarios \citep{xu2023multi}.

Moreover, this MAS approach has relatively low adoption in industries \citep{marik2020multi, karnouskos2020industrial}.
As discussed by \citet{karnouskos2020industrial}, aside from business-relevant factors, the lack of terminologies and standards contributed to this limited acceptance.

This is particularly evident in the emerging domain of ASC, which involves a network of cooperative yet competitive distributed stakeholders. 
Therefore, a generalised conceptual framework containing essential elements such as definitions, terminologies, and technical maturity levels is crucial for both the conceptual and technical development of ASC.

\section{The Autonomous Supply Chain Model}\label{sec:asc}
Aside from introducing preliminaries, this section describes key concepts for conceptualising the ASC model, including its definition and characterising features.

\subsection{Preliminaries}\label{sec:preliminaries}
\begin{figure}
    \centering
    \includegraphics[width=0.85\textwidth]{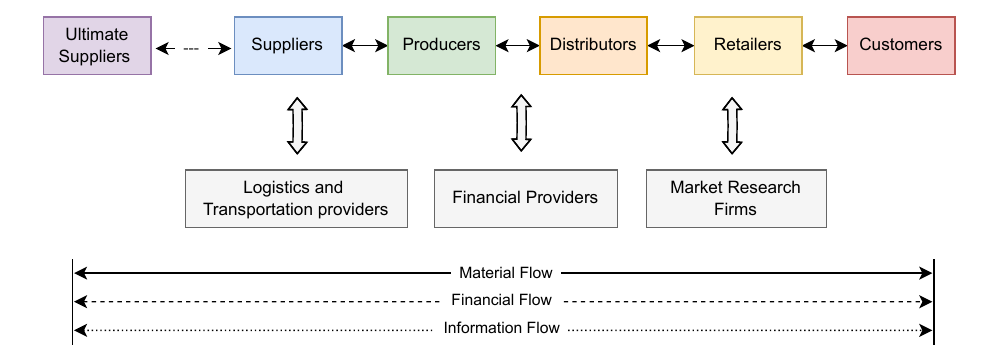}
    \caption{Illustration of an extended supply chain.}
    \label{fig:sc_illustration}
\end{figure}
Before dealing with the ASC model, we first briefly describe relevant preliminaries: networked supply chains and SCM.

\subsubsection{Networked Supply Chains}\label{subsec:sc}
A supply chain is a set of business entities involved in the movement of materials, services, information, and finance from sources to customers, both upstream and downstream \citep{mentzer2001defining, christopher2016logistics}. 
\autoref{fig:sc_illustration} illustrates an extended supply chain, with entities that constitute the {\it static} aspects of a supply chain, along with additional service providers such as third-party logistics (3PL) and financial providers that offer specific services to the stakeholders in the supply chain. 
The {\it dynamic} aspects of a supply chain are represented by three key flows: material flow, financial flow, and information flow, forming a complex adaptive network \citep{choi2001supply, proselkov2024financial}.

The supply chain depicted in \autoref{fig:sc_illustration} is visually simplified as a linear structure and includes only a subset of entities for clarity.
However, it is important to note that modern supply chains have evolved into more complex and large-scale structures characterised by interdependence and connectivity, forming a {\it networked} structure \citep{brintrup2015supply}.

\subsubsection{Supply Chain Management}\label{subsec:scm}
Effective supply chain design and management is crucial to every company \citep{stevens1989integrating}. 
SCM forms the backbone of most economies and successful multinationals today.
The significance of managing supply chains can be, at least, traced back to the creation of the assembly line in the early 20th century.
However, the term ``supply chain management'' was coined only a few decades ago by Keith Oliver in a 1982 interview with The Financial Times \citep{oliver1982supply}, and gained prominence in the late 1990s.
Since then, various definitions of SCM have been proposed, each with a different focus, evolving to align with changing business environments and technology advancements \citep{stevens1989integrating, cooper1997supply, mentzer2001defining, christopher2016logistics}.

To reduce confusion and ambiguity, \citet{mentzer2001defining} examined prior SCM concepts and definitions and proposed a consistent means to conceptualise SCM.
They proposed the concept of supply chain orientation (SCO), which represents the management philosophy that organisations must view SCM activities and coordination from a {\it systemic} and {\it strategic} perspective.
This SCO then forms the basis for SCM, which is defined as the coordinated set of inter-firm and intra-firm actions implemented to embody this philosophy \citep{mentzer2001defining}. 
By demarcating SCO from SCM and positing SCO as a major antecedent of SCM, this definition provides a  holistic, strategic, and broader view on SCM conceptualisation \citep{min2019defining}.
This influential definition highlights inter-functional and inter-corporate coordination, aligning with the consistent claim about the strategic importance of integrating upstream and downstream supply chains \citep{stevens1989integrating, frohlich2001arcs, childerhouse2011arcs, christopher2016logistics}.

Market environments and technologies have undergone significant changes since the development of these concepts nearly two decades ago.
As described in \autoref{sec:introduction}, business environments have evolved to become more volatile and customer-centric. 
Additionally, new advanced technologies such as AI, IoT, and advanced robotics have emerged and are now maturing. 
These technologies are transforming SCM, automating numerous supply chain functions.
Recently, an article by \citet{lyall2018death} even asserted `the death of SCM', arguing that digital technologies are making traditional SCM obsolete.

However, a closer examination offers an alternative interpretation: while the implementation methods are changing, the core elements of SCM --- their strategic nature, customer value creation, and inter-organisational collaboration --- remain pertinent today \citep{stevens2016integrating, christopher2016logistics, min2019defining}.
Rather than outdating SCM, technologies are disrupting its practices, paving the way for new SCM models.
The ASC proposed in this paper represents such a technology-enabled SCM model, built upon the foundational elements that have long been recognised within SCM.

\subsection{Defining the ASC: Type of Connections, Structural Entities, and Its Definition}
\label{sec:definition}
The ASC concept or similar ideas have been discussed in both academic literature (e.g., \citet{wu2016smart, calatayud2019self}) and industrial reports (\citet{lyall2018death, blueyonder2020autonomous, nelsonhall2021moving}, etc.) over the past few years, mostly as derivative of the advancement of AI, IoT and/or advanced robotics.   
While these interpretations of the ASC vary, they are commonly characterised by technology-enabled features, including automated processes, continuous monitoring, and unmanned decision-making.
It is worth noting that, as of now, there is no well-defined definition of ASC in the literature. 
Therefore, this section provides a definition of the ASC and describes its defining characteristics and relevant concepts.

Just as with autonomous vehicles (AVs), the most crucial feature of an ASC is {\it autonomy}. 
The term ``autonomy'' is defined by the Merriam-Webster dictionary as ``the quality or state of being self-governing'' \footnotetext{\url{https://www.merriam-webster.com/dictionary/autonomy}}. 
The key aspect here is ``self-governing'' --- the capability to determine, conduct and control the actions or behaviours independently, without external input. 
When applied to vehicles, autonomy refers to the vehicle's capacity to self-govern its driving. 
Similarly, in supply chains, autonomy implies the ability to self-govern its operations. 
However, the distributed and decentralised nature of ASCs differentiates then from  monolithic autonomous systems like AVs.
Monolithic systems operate within a singular organisational scope without the need to coordinate with other self-governing entities.
Most existing autonomous systems are monolithic. 
For example, an AV achieves autonomy by orchestrating the functions of its diverse components either within the vehicle or through remote servers, involving a central governing entity.

A supply chain is a network of interconnected entities, each of which may have independent decision-making capability. 
This network is established with the goal of making products available to meet customer demand. 
To achieve this goal, a supply chain requires coordination among the various entities that constitute it, as well as coordination among the components within each individual entity. 
Whether self-organised, intentionally designed, or a combination of both, supply chains exhibit varying degrees of connections. 
This includes the connections between the entities themselves and between the internal components within each entity.
We classify these connections into two types: {\it external} connections and {\it internal} connections. 
In external connections, we further identify two classes of relationships based on the degree of coupling \citep{brown2002loosening} between supply chain entities: {\it tight} and {\it loose} external connections, as described by:
\begin{itemize}
  \item {\it Tight External Connection}: This denotes a close relationship between two entities, characterised by high interdependency. 
  Such connections are often intentionally designed and can be considered ``hard-wired'' due to their strong and deliberate linkage.
  \item {\it Loose External Connection}: This refers to an arm's length relationship between two entities, characterised by low interdependency.
  These connections offer higher flexibility, as the entities involved maintain a higher degree of independence.
\end{itemize}

\begin{figure}
    \centering
    \includegraphics[width=0.90\textwidth]{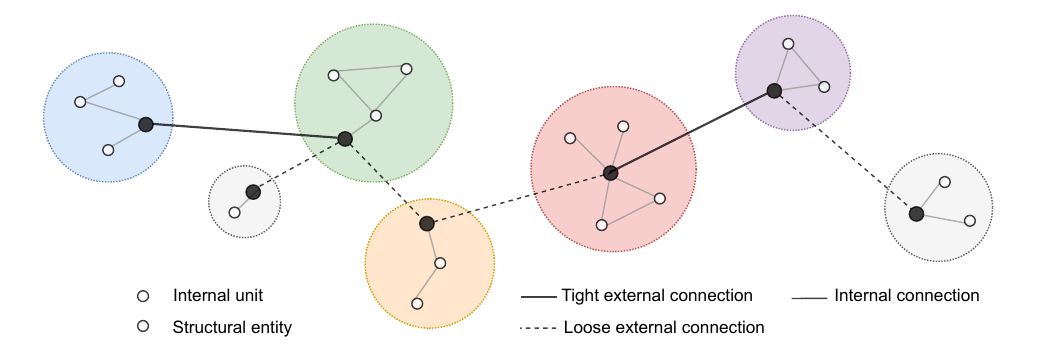}
    \caption{Illustration of the ASC structure.}
    \label{fig:structural_entity}
\end{figure}
Tight and loose external connections allow companies to be linked either through predefined configurations or through emergent arrangements when structuring supply chains.
Unlike external connections, which require the establishment of common rules or principles to facilitate interaction, internal connections are intrinsic to the organisation of each individual company and therefore under the same span of control. 
Internal connections describe the interdependency between units within an entity, where all units coordinate coherently towards a common objectives.
These connections are defined by the company itself and reflect its internal organisational structure and dynamics.

To ensure the efficient functioning of an ASC, close collaboration among all supply chain stakeholders is essential. 
These stakeholders include both the entities within the supply chain and their internal operational divisions.
We introduce a new concept, the Structural Entity, representing the local authority at an external connection point. 
Structural entities are defined as follows:
\begin{displayquote}
    A {\it structural entity} is the entity along the supply chain network that assembles and controls the flows of materials, information and finance.
    Structural entities gather the essential data needed by other entities during decision-making. 
\end{displayquote}

Structural entities compose the main structure of an ASC, bridging external and internal connections, enabling entities of varying autonomy levels to interact.
These entities often represent firms or groups of supply chain entities with shared interests.
We illustrate an ASC composed of structural entities and internal units in \autoref{fig:structural_entity}, in which structural entities and internal units are denoted by black-filled circles and unfilled circles, respectively.
As shown in \autoref{fig:structural_entity}, the supply chain consists of a collection of tightly or loosely connected organisations denoted by grey circles, each composed of a set of internal units.
Regardless of whether these organisations are autonomous or not, all structural entities in an ASC must operate coherently without human intervention. 
Based on the concepts we presented, we define an ASC as follows: 
\begin{displayquote}
    An {\it autonomous supply chain} (ASC) is a self-governing supply chain built upon intelligence and automation, in which key structural entities are capable of making and enforcing their decisions with little or no human intervention.
\end{displayquote}

As presented in the definition above, intelligence and automation are fundamental elements of an ASC. 
Intelligence refers to the ability to make decisions and develop solutions to problems in a dynamic and uncertain environment \citep{gunderson2004intelligence}, while automation involves the capability to execute solutions automatically. 
These two aspects form the two dimensions of the autonomy manifold, as shown in \autoref{fig:autonomy_manifold}. 
Further details regarding this manifold are discussed in \autoref{sec:autonomy_manifold}.

As a distributed and decentralised autonomous system, an ASC comprises a set of key structural entities that are autonomous and inclined to coordinate with each other for mutual benefit. 
Some of these structural entities may be grouped due to the scale and performance needs of the supply chain.
Therefore, at least some representative structural entities must possess autonomy.
These autonomous structural entities function as {\it connection points}, managing materials, information, and financial flows.
To ensure smooth operation of an ASC, well-defined communication and interaction mechanisms among entities must be in place. 
These mechanisms ensure both entity autonomy and coordinated behaviour across all entities. 
They specify how and what entities communicate, including languages, protocols, and terminologies for effective communication and interaction.

\subsection{Characterising the ASC: Two Dimensional Autonomy and Six Characteristics}
\label{sec:characteristics}
The previous section presents the definition of an ASC, based on the factors of intelligence and automation. 
This section examines the manifold formed by these two factors and subsequently defines the characteristics of ASCs.

\subsubsection{Two Dimensional Autonomy}\label{sec:autonomy_manifold}
\begin{figure}
    \centering
    \includegraphics[width=0.50\textwidth]{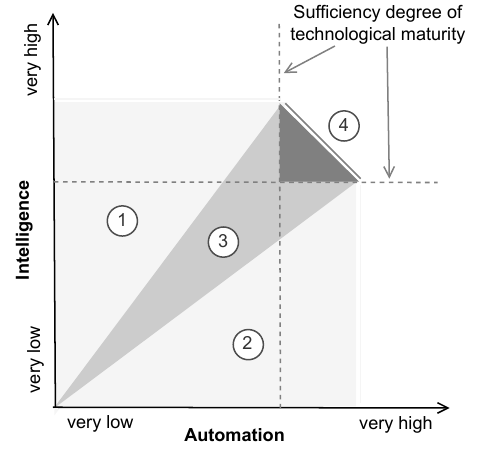}
    \caption{The two-dimensional autonomy manifold. 
             The labelled circles, numbered 1 to 4, represent the four regions: intelligence-skewed, automation-skewed, balanced, and ideal.}
    \label{fig:autonomy_manifold}
\end{figure}
As discussed in \autoref{sec:definition}, autonomy has two dimensions: intelligence and automation. 
These two dimensions form a bounded space within which the autonomy of a supply chain entity can be evaluated. 
The extent of each dimension is constrained by the current level of technological maturity, but there is potential for expansion through technological advancement.

The resulting autonomy manifold is shown in \autoref{fig:autonomy_manifold}.
As depicted, the autonomy manifold can be divided into four regions: 
intelligence-skewed, 
automation-skewed, 
balanced, and 
ideal. 
These regions are annotated with circled numbers from one to four. 
Dashed horizontal and vertical lines indicate the {\it sufficiency degree of technological maturity} in their respective dimensions.
Entities located within the skewed regions have achieved greater technological development in one dimension than the other, resulting in a {\it compromised} level of autonomy. 
For instance, an intelligence-skewed entity can automatically propose solutions but might struggle to effectively execute these solutions without human assistance. 
Conversely, an entity skewed towards automation possesses the capability to perform planned actions but may lack the ability to generate appropriate actionable plans. 
In an ASC, entities, particularly structural entities, should strive to position themselves in the spectrum of dark regions (the diagonal area in \autoref{fig:autonomy_manifold}), where they achieve a relative balance between intelligence and automation.
The darkest region (denoted by circled four) represents the ideal region, situated above both technological maturity lines. 
Entities in this region attain {\it sufficient autonomy}, with well-balanced and sufficient unmanned decision-making and execution capability. 
Importantly, within the balanced and ideal regions, the two dimensions may not necessarily be equal;  
one dimension might outpace the other within a certain range and time frame, allowing entities to take measures to rebalance and enhance their autonomy.

\subsubsection{The Six Characteristics}\label{sec:six_characteristics}
The core defining characteristic of an ASC is {\it autonomy}, which is further characterised by the two dimensions depicted in \autoref{fig:autonomy_manifold}.
Previous studies, such as \citep{butner2010smarter, wu2016smart, calatayud2019self}, envisaged future supply chain models and summarised their features, as described in \autoref{sec:related_work}.
These characteristics often focus on technology-enabled aspects such as continuous monitoring, data connectivity, process integration and automation, and predictive analytics. 
While these features are critical for the operating of an ASC, they were initially conceptualised in response to the development of corresponding emerging technologies and were often treated as isolated concepts, without fully considering their interrelationships.

We herein propose six bottom-up layered characteristics for defining ASCs: 
instrumented, 
standardised, 
interconnected, 
integrated, 
automated, and 
intelligent. 
These six characteristics are defined without being tied to specific implementation technologies.
Furthermore, they are logically separate but functionally connected.
These characteristics are:
\begin{enumerate}
  \item[1)] {\it Instrumented}: 
  Data connectivity is facilitated through installed instruments.
  Various devices (sensors, actuators, tags, etc) are employed by supply chain entities for real-time data collection and transmission, tracking, monitoring, and analytics.
  
  \item[2)] {\it Standardised}: 
  Common standards and rules govern processes related to inter-entity interaction.
  Procedures, standards, protocols, guidelines, regulatory frameworks, and data formats for data exchange between entities are established.
  
  \item[3)] {\it Interconnected}: 
  Supply chain entities, internal units within entities, and other objects and systems that support supply chain operations are connected. 
  This extensive interconnection allows entities to interact with others under certain guidelines and/or protocols.
  
  \item[4)]  {\it Integrated}: 
  A broad {\it arc of integration} is implemented, enabling entities to coordinate with upstream and/or downstream counterparts in shared operational activities. 
  
  \item[5a)] {\it Automated}:
  Automation is enabled and leveraged. 
  Workflows are designed to allow machines to perform efficiently and effectively, and tasks and processes in the supply chains are automated. 
  
  \item[5b)] {\it Intelligent}: 
  Key entities possess autonomous decision-making capabilities in the function they are tasked to manage. 
  Entities can reason based on their context and propose suitable actions with minimal human intervention. 
\end{enumerate}

While some of the characteristics presented above have been introduced in prior work such as \citet{butner2010smarter, wu2016smart}, they have been adopted and redefined to suit the context of ASCs. 
We organise these characteristics into a bottom-up layered structure to provide a framework for characterising ASCs.
They are not reliant on specific technologies; rather, they focus on describing how technological advancements enhance and enrich ASC implementations.
The first four characteristics are downward dependent, with higher-layer characteristics facilitated by the lower ones.
In contrast, the final two characteristics, automated and intelligent, are relatively less interdependent; they are manifest features of an ASC.

\section{A Conceptual Framework: the MIISI Model}\label{sec:miisi_model}
In this section, we present a layered conceptual model --- the MIISI model --- for constructing ASCs, detailing the main functions and data associated with each layer within our proposed model.

\begin{figure}
    \centering
    \includegraphics[width=\textwidth]{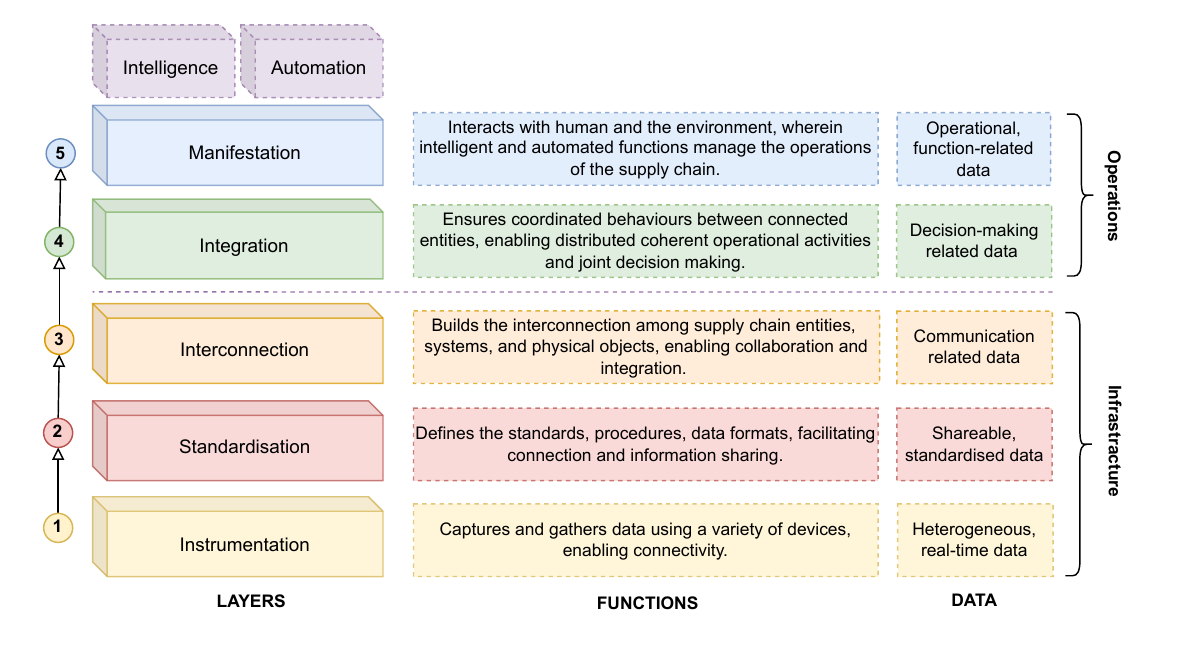}
    \caption{Illustration of the proposed MIISI model.}
    \label{fig:miisi_model}
\end{figure}

The previously defined bottom-up, layered six characteristics suggest potential conceptual partitions for a functional ASC system that encompasses diverse entities with varying degree of autonomy.
It is thus natural to translate the descriptions of each characteristic into corresponding abstraction layers.
The two characteristics, ``intelligent'' and ``automated'', have a relative independence from each other and directly contribute to the manifestation of an ASC.
Therefore, they are grouped into a single layer termed the ``Manifestation'' layer.
The remaining characteristics each relates to a separate layer, collectively facilitating the manifestation layer.
This partition results in a five-layer conceptual model, each building upon the previous one. 
As illustrated in \autoref{fig:miisi_model}, these layers from top to bottom are: 
Manifestation,
Integration, 
Interconnection, 
Standardisation, 
Instrumentation.
This conceptual model is referred to as the MIISI model, an acronym for the five layers in top-down order.

Each layer in the MIISI model has distinct overarching functions, each building upon the layers beneath it. 
The MIISI model assigns different levels of abstraction and scope of view to each layer, with higher layers being more abstract and having a broader scope of view. 
The bottom layer deals with concrete objects and has a local view, whereas the top layer has an integrated, global view of the supply chain, focusing on specific operational processes.
The middle layers facilitate bridging the bottom and top layers by standardising data flow and processes and establishing connections with entities.

The main functions and the data handled in each layer are detailed as follows:
\begin{enumerate}[nosep]
    \item\label{lyr:instrumentation} {\it Instrumentation Layer}: 
    This layer is responsible for collecting and aggregating raw data necessary to monitor the environment and establish connectivity. 
    It involves utilising a diverse array of equipment, devices, protocols, and systems to generate, capture, store, and transmit data. 
    This layer also specifies the scheme and metadata of any attributes (type, format, variety, quality, purpose, and owner, etc.) of objects needed to enable automated functions or processes.
    Furthermore, it supports two types of connectivity: 
    \begin{itemize}
        \item Local Connectivity: Enables communication between physical and/or digital things within an organisation. 
        
        \item Wide Connectivity: Allows connections to external networks owned by other entities. 
    \end{itemize}
    Both types of connectivity support intra- and inter-company information sharing, which is a critical aspect for enabling ASC functionalities. 
    
    \vspace{5pt}
    
    \item\label{lyr:standardisation} {\it Standardisation Layer}: 
    This layer is responsible for setting standards and establishing procedures for data sharing and exchange, thus enabling process automation.
    It defines data schemes, data interchange formats, and agreed-upon procedures to facilitate automated data exchange between entities involved in tasks. 
    Two types of data exchange are considered: {\it internal} data exchange within an organisation and {\it external} data exchange with entities outside the organisation.
    Data for each type of exchange needs to be defined in a standardised, machine-readable format to enable automated data manipulation and interpretation. 
    However, entities may have different data sharing policies regarding data security and privacy. 
    This layer identifies processes needed to be standardised for interconnection and interoperability, defining agreed-upon procedures for performing processes related to the three flows. 
    Standardised processes and data exchange procedures facilitate the connection and coordination between entities. 
    
    \vspace{5pt}

    \item\label{itm:interconnection} {\it Interconnection Layer}: 
    This layer is responsible for managing connections between entities, physical objects, or digital objects across the supply chain.
    It identifies entities, establishes, manages, and terminates connections between two or more parties. 
    Key functionalities of this layer include registration and naming services, authentication, message delivery, and error detection and correction.
    Utilisation of shared infrastructure services might be necessary for implementing registration, naming, and reliable message delivery.
    Protocols are also established within this layer to regulate connection.
    This layer facilitates the connection of entities, physical objects, and digital objects across the supply chain as required, allowing {\it ad hoc} or {\it dynamic} configuration of supply chain networks. 
    
    \vspace{5pt}
    
    This layer considers three types of connections: {\it ad hoc}, {\it temporary}, and {\it established}. 
    These three connections are categorised based on the purpose of the connection and the relationship between the connected parties, detailed as below: 
    \begin{itemize}
        \item Ad Hoc Connection: This involves connecting two unknown parties to address specific needs or problems, whether for short or long-term purposes.
        These connections are often spontaneous and solution-focused.
        
        \item Temporary Connection: This refers to the connection between two unrecognised parties, but they connect only for short-term purposes.
        These connections are typically formed for specific projects or short-term collaborations.
        
        \item Established Connection: This involves mutually trusted parties that may have an established contract or a long-standing collaboration. 
        These connections are stable and ongoing, often involving formal agreements and long-term cooperation.
    \end{itemize}
    Various design considerations, including authentication, connection modes, data security and privacy, among others, may be necessary for these different types of connections.
    This layer facilitates the {\it customised} formation of supply chain structures and enables coordinated behaviour and joint decision-making among participants, thereby supporting the implementation of various integration strategies. 
    
    \vspace{5pt}
    
    \item\label{itm:integration} {\it Integration Layer}: 
    This layer coordinates and aligns the behaviours of interconnected entities in the supply chain, facilitating collaboration within individual entities and between various participants in the supply chain, with the ultimate goal of enabling them to operate cohesively as a unified whole.
    It ensures that decentralised and geographically dispersed stakeholders collaborate effectively towards common goals that benefit the entire supply chain.
    Moreover, it supports different types of integration with varying breadths, ranging from the narrowest {\it inward-facing} integration to the broadest {\it outward-facing} integration \citep{frohlich2001arcs}. 
    To achieve this, this layer needs to establish protocols for negotiation, define agreed-upon procedures for optimising collaborative processes, and provide standardised methods for preparing shareable data (such as data encryption, compression, and conversion) to enable a wide range of integration strategies. 
    Additionally, this layer converts data into a format suitable for the layer immediately above it.
    Leveraging the capabilities of the interconnection layer, this layer ensures collaborative decision-making and enables coherent operational activities among entities.
    
    \vspace{5pt}
    
    \item\label{itm:manifestation} {\it Manifestation Layer}: 
    This layer is responsible for managing the day-to-day operations of an ASC.
    It includes a set of intelligent and automated applications, devices, and machinery that interact with humans and perform operations within the supply chain.
    However, this layer is not designed to take over every business process or activity.
    Instead, its primary focus is on managing the automated movement of three key elements: information, finance, and product.
    Two categories of flow are defined in this layer based on the conveyance manner: {\it digital} and {\it physical}.
    Digital flow relates to intangible items that can be represented in the form of {\it bits}, while physical flow involves tangible objects that exist in the form of {\it atoms}. 
    Consequently, the flows of information and finance are considered digital flows, as they can be represented by bits and transmitted along the supply chain using digital communication methods such as the Internet. 
    On the other hand, the movement of tangible products constitutes a physical flow, where their movement needs to be completed through transportation systems.
    It is important to note that intangible products, which are not physical in nature and often exist digitally (e.g., licenses and software), can also be represented as bits and moved through the Internet.
    Therefore, their movement is considered as a special form of digital flow within this layer. 

    \vspace{5pt}

    This layer comprises a range of self-regulating utilities that embody the dimensions of autonomy described earlier (see \autoref{sec:autonomy_manifold}).
    These utilities have the capability to make decisions autonomously, including managing supply chain planning across various time horizons, proposing real-time solutions for emergent events, and predicting and responding to contingent events.
    Additionally, they can execute courses of action to implement decisions, such as rerouting to avoid traffic congestion, replenishing inventory, and selecting alternative suppliers.
    Entities equipped with intelligent and automated systems can then collaborate to plan, control, and execute the flow of both digital bits and physical atoms along the supply chain.
    This layer mainly handles operational and application-related data.
    
\end{enumerate}

\vspace{5pt}

Digital flows, such as information and financial flows, are often considered to be driven by physical flows and therefore secondary. 
However, we argue that ASCs should treat digital flows with equal importance to physical flows, ensuring the efficient and seamless movement of data across the supply chain.
Prioritising digital flows does not undermine the importance of physical flows; instead, it enhances the management of physical flows. 
Even if physical flows cannot be entirely automated and requires manual interventions, human involvement can align it with the fully automated digital flow, thus addressing issues related to human errors and inefficiencies in moving data.

This model stratifies an ASC into five abstract and conceptual layers. 
The lower three layers form the foundational infrastructure of ASCs, enabling connections and facilitating coordination among supply chain entities.
The top two layers are focused on the everyday management and operation of ASCs, involving applications responsible for various functions, with minimal or no human intervention. 
This conceptual model describes an unmanned SCM system but does not necessarily imply the complete replacement of human workers with machines in the ASC era.
While automation may reduce the numbers of frontline human workers \citep{hawksworth2018will}, it also creates opportunities for new types of jobs or ``job of tomorrow'' \citep{orduna2021why}.
Machines will play a crucial role in ASCs, but humans will remain integral.
Their role will shift from mundane and repetitive tasks to more strategic activities, such as guiding machines in supplier selection. 
Achieving full supply chain autonomy is a complex journey that requires both technological advances and managerial efforts.
The next section discusses the stages towards realising a fully autonomous supply chain.

\section{The Seven Supply Chain Autonomy Levels}\label{sec:autonomy_level}
Achieving supply chain autonomy is not an {\it all-or-nothing} proposition; rather, it is a {\it gradual process that unfolds over time}.
Companies progressively integrate automated functionalities into their supply chains.
A maturity reference model outlining the stages of supply chain autonomy development would benefit all stakeholders by providing a shared framework for discussing the technical development of supply chain autonomy within both academic and industrial spheres.
Additionally, it can be used as a benchmark for comparison, aiding executives in making informed decisions.
However, such models are currently lacking in both academia and industry.
This section seeks to bridge this gap by introducing a consistent framework for assessing the level of supply chain autonomy.

Several models exist for evaluating technological developments.
The Technology Readiness Levels (TRLs) and the Capacity Maturity Model (CMM) \citep{humphrey1988characterizing, paulk1993capability} are two widely recognised frameworks.
The TRLs, originating from NASA in the 1970s, include a nine-level scale used to assess the maturity of specific technologies.
The CMM serves as a process maturity framework, delineating five levels to measure the maturity of an organisation's software processes.
Both models provide an incremental path for continuous improvement and have been gained adoption across various industries.

Additionally, models exist for appraising the developmental stages of autonomous systems.
\citet{dumitrescu2018studie} outlined five development stages of technical systems, such as robots, vehicles, machines and software, leading toward autonomous systems.
For a concise review in English, one can refer to \citet{nitsche2021exploring}.
Notably, the Society of Automotive Engineers (SAE) developed a classification system \citep{sea2021taxonomy} that includes six levels of vehicle autonomy, ranging from 0 (fully manual) to 5 (fully autonomous). 
This system has gained recognition as a standard for assessing the degree of driving automation in vehicles.
Other works include the three levels of object intelligence \citep{brintrup2011will} and the three phrases for implementing a smart supply chain \citep{wu2016smart}.

\begin{figure}[t]
    \centering
    \includegraphics[width=\textwidth]{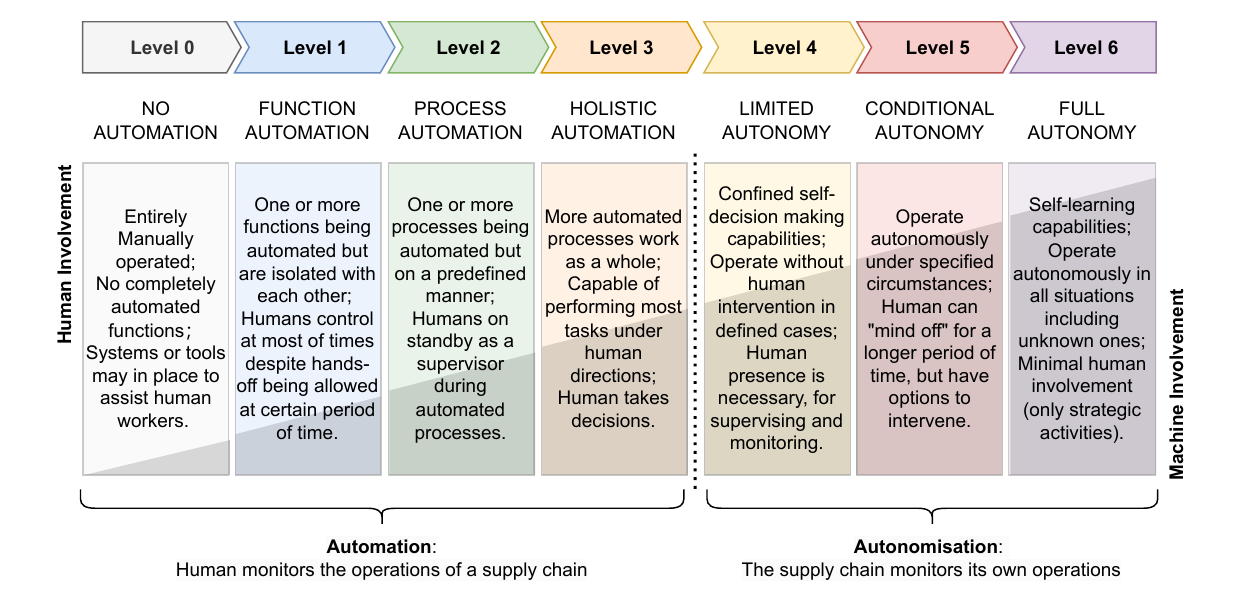}
    \caption{The seven supply chain autonomy levels (SCALs).}
    \label{fig:autonomy_levels}
\end{figure}

Drawing inspiration from these models, particularly the SAE driving automation levels, we define seven distinct levels for evaluating supply chain autonomy.
These levels, called Supply Chain Autonomy Levels (SCALs), as shown in \autoref{fig:autonomy_levels}, range from level 0 (L0: fully manual) to 6 (L6: fully autonomous), providing a phased process toward full supply chain autonomy.
The proposed seven SCALs are detailed as follows: 
\begin{enumerate}[nosep, label=L\arabic*]\setcounter{enumi}{-1} 
    \item\label{itm:l0} {\it No Automation}:
    supply chains at this level are operated entirely manually. 
    Human operators are solely responsible for managing and performing all aspects of the supply chain operations. 
    Although supportive systems or tools may be in use, human intervention remains constant throughout the process.

    \vspace{5pt}
    
    \item\label{itm:l1} {\it Function Automation}: 
    At this level, certain functions within the supply chain have been automated. 
    However, these automated functions are often disconnected from each other and do not form a continuous process.
    Human operators are responsible for connecting manual and automated functions throughout the process.
    While humans may be on standby during the execution of certain automated functions, their active involvement is still prevalent, and they retain overall control over supply chain operations.

    \vspace{5pt}
    
    \item\label{itm:l2} {\it Process Automation}:
    At this level, one or more processes in the supply chain are automated and streamlined.
    A process comprises a sequence of functions that are executed in a predetermined order.
    Although hands-off operations are allowed during automated processes, human operators must remain vigilant and ready to intervene upon request. 
    The key distinction between L2 and L1 is the integration of multiple automated functions, creating a connected series of actions that automatically execute a specific process. 
    Supply chains at this level can perform automated processes and respond precisely to {\it predefined} situations.
    These supply chains exhibit fewer human errors, enhanced efficiency, and reduced reliance on frontline operators. 
    However, additional personnel may be required to oversee the execution of automated processes. 
    While humans maintain overall control of supply chain operations, their role primarily involves supervision as they standby during the execution of processes for the majority of the time.

    \vspace{5pt}
    
    \item\label{itm:l3} {\it Holistic Automation}: 
    When the operation of a supply chain becomes nearly entirely automated, it progresses to L3.
    At this level, all major processes within the supply chain are automated and connected to work coherently with human intervention. 
    Compared to L2, an L3 supply chain encompasses a greater number of automated processes, even involving actions that require external entities' participation. 
    Automation permeates nearly every aspect of the supply chain, but it primarily focuses on executing operational decisions rather than automating decision making.
    Supply chains at this level are highly automated and capable of automatically performing the majority of tasks under human guidance.
    The need for front workers is further reduced, although more individuals may be involved in supervising and monitoring the execution of automated processes.
    Humans still control decision making, particularly on tactical and strategic levels.

    \vspace{5pt}
    
    \item\label{itm:l4} {\it Limited Autonomy}:
    The transition from L3 to L4 signifies a substantial technological advancement, although it may appear subtle or negligible from a human perspective. 
    The defining characteristic of an L4 supply chain is its capability for self-decision-making.
    Specifically, the supply chain can autonomously perceive its environment and make decisions based on its accumulated knowledge, without requiring human input.
    However, this self-decision-making is confined to specific functions, human intervention is still essential beyond these functions.
    For example, the supply chain can decide to replenish inventory without human instruction in anticipation of an upcoming demand surge, but it may need human guidance in selecting appropriate suppliers.
    At L4, the supply chain shows a considerable level of automation and begins to acquire limited autonomy, with a limited self-learning capability to expand its knowledge base.
    Although human involvement is still necessary, it mostly revolves around making decisions in complex situations and monitoring the execution of automated processes.

    \vspace{5pt}

    \item\label{itm:l5} {\it Conditional Autonomy}: 
    A supply chain reaches L5 when it can perform all functions autonomously without human intervention under certain conditions. 
    This supply chain can operate independently in predefined circumstances, while humans remain alert to respond to any requests for intervention in unexpected events.
    At this level, humans maintain a high-level control over the supply chain operations and can be in a state of ``mind off'' for extended periods.

    \vspace{5pt}

    \item\label{itm:l6} {\it Full Autonomy}: 
    An L6 supply chain achieve complete automation, possessing full self-learning and self-decision-making capabilities, and can operate autonomously for extended periods.
    This supply chain can handle all situations with minimal or even zero human attention or interaction, even in unanticipated situations.
    At this level, human involvement is kept to a {\it minimal} level.  
    Human focus may shift to strategic activities that are not necessarily directly involved in the everyday operation of the supply chain.
    
\end{enumerate}

These seven levels represent a trajectory leading towards full supply chain autonomy (see \autoref{fig:autonomy_levels} for an illustration of these levels).
This trajectory comprises two distinct phases: {\it automation}, encompassing the initial four levels (L1 -- L4), and {\it autonomisation}, which includes the three higher levels (L4 -- L7) in the SCAL reference model.
As the level of autonomy increases, the supply chain incorporates more automated functions and acquires self-decision-making capabilities, becoming increasingly capable of handling more complex scenarios with less human intervention.
Throughout this trajectory, human involvement decreases while machine involvement increases, as illustrated by the shrinking light grey area and expanding dark grey area in \autoref{fig:autonomy_levels},
While the differentiation between consecutive levels within each phrase is based on the degree of automation, the distinction between the two phrases is substantial.
During the automation phase, humans mainly operate the supply chain, whereas in the autonomisation phase, the supply chain starts to manage its own operations.
Exactly assigning an existing supply chain to a specific level may not always be possible due to complexity of real-world supply chains. 
This spectrum of supply chain autonomy provides a lens through which companies can examine the current stages of automation development in their supply chains and guide their automation strategies.

\section{A Case Study: Autonomous Meat Supply Chain}\label{sec:case_study}
\begin{figure}
    \centering
    \includegraphics[width=0.725\textwidth]{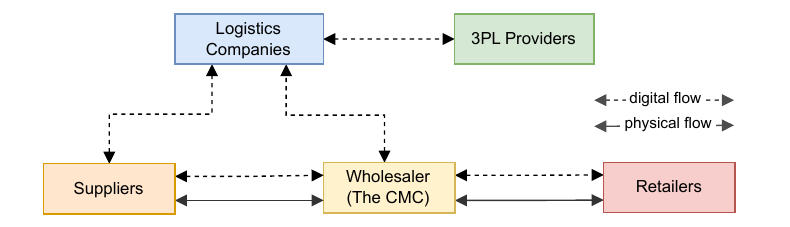}
    \caption{The physical and digital flows of the meat supply chain under study.}
    \label{fig:case_flows}
\end{figure}
In this section, we provide a brief overview of an ASC prototype implementation, demonstrated through a case study involving the development of a platform for automated meat procurement and wholesale.
For reasons of brevity, we exclude technical details and focus on demonstrating how the proposed conceptual model---the MIISI model---guides the design and development of a concrete ASC. 
More details of this system, including its design and technical implementations, can be found in \citet{xu2024implementing}.

\subsection{Design}\label{sec:design}
This particular application involves a local meat company called the Cambridge Meat Company (CMC).
The CMC specialises in the wholesale procurement of meat, such as chicken, beef, and lamb, and its subsequent distribution to local restaurants.
It is important to emphasise that the CMC is a {\it hypothetical} company created solely for illustrative purposes.
Its supply chain includes several key participants: 

\vspace{5pt}

\begin{itemize}[nosep]
    \item {\it Suppliers}: Entities such as meat producers or suppliers who provide meat products. 
    \item {\it Wholesaler}: The CMC itself, acting as an intermediary that purchases meat from suppliers and sells it to retailers. 
    \item {\it Retailers}: Businesses such restaurants or local stores that purchase meat products from the CMC. 
    \item {\it Logistics Companies}: Companies responsible for managing the logistics of transporting goods, specifically meat, within the supply chain. 
    \item {\it 3PL Providers}: External logistics service companies that offer delivery services to logistics companies, transporting meat products from their source to assigned destination. 
\end{itemize}

\vspace{5pt}

This supply chain consists of two main processes: {\it replenishment}, where the CMC procures meat from suppliers to restock its inventory, and {\it wholesale}, where the CMC acts as a wholesaler, providing meat products to retailers, such as local restaurants.
Both processes need logistics services for order fulfilment, with the sellers responsible for handling delivery arrangements.

\begin{figure}
    \centering
    \includegraphics[width=\textwidth]{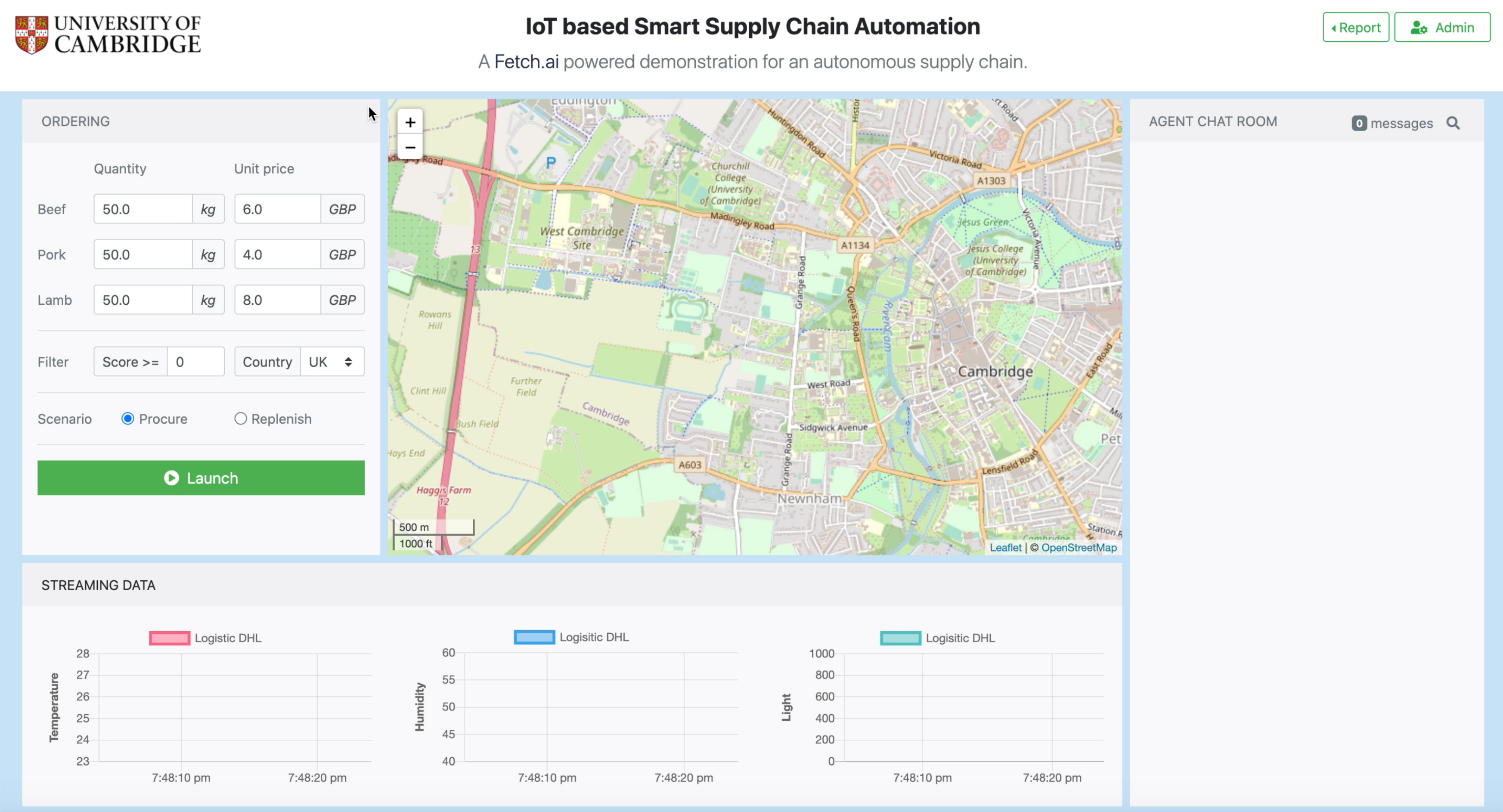}
    \caption{A screenshot of the system interface in its startup state, with all four panels are in their initial state with empty content and default values.}
    \label{fig:interface_startup}
\end{figure}

To achieve full automation of these two processes, multiple decisions must be made autonomously.
These decisions include creating proposals, accepting or declining proposals, and selecting appropriate delivery options.
In real-world scenarios, these decision-making processes are often highly complex, requiring one to consider numerous short- and long-term factors. 
For illustration purposes, we simplify these decisions by only considering a set of simple predefined rules. 
Using these simplified settings, we implemented an autonomous meat supply chain system (see \autoref{fig:interface_startup}) following the MIISI model introduced in \autoref{sec:miisi_model}.

This autonomous meat supply chain is {\it simulated}, with no actual purchases and transportation occurring. 
This supply chain is shown in \autoref{fig:case_flows}, where its two types of flows are represented by dashed arrows and solid arrows, denoting potential interactions between stakeholders.
To develop a system capable of automating meat procurement and wholesale, we employ a MAS approach \citep{wooldridge1995intelligent, fox2001agent}, which is well-suited for connecting distributed entities and thus facilitating supply chain integration. 
Due to space limitation, we omit the implementation details in this paper, as they are beyond its scope.

The resulting platform is a web-based system accessible via a web browser. 
As shown in \autoref{fig:interface_startup}, the system's interface consists of four panels: ordering (top left), transport monitoring (central), agent chat room (right), and product ambient condition monitoring (bottom). 
These panels correspond to the three basic procedures in a supply chain: 

\begin{itemize}[nosep]
    \item {\it Ordering}: This panel handles purchasing-related functions such as supplier selection, order placing, and order confirmation. 
    The ordering panel is the interface for ordering. 
    
    \item {\it Logistics Monitoring}: This pertains to shipping goods, tracking their locations, and monitoring ambient conditions. 
    This process is depicted in the central and bottom panels of the interface. 
    
    \item {\it Negotiation and Coordination}: Supply chain stakeholders interact with one another to resolve disagreements and collaborate effectively, as shown in the agent chat room panel.
\end{itemize}

\vspace{5pt}

Logistics occurs after the ordering process, and negotiation and coordination are integrated into these two processes to resolve conflicts and achieve coherent behaviours.
In this system, a set of ``autonomous'' agents is developed to act on behalf of these stakeholders shown in \autoref{fig:case_flows}.
These agents collectively manage the operations of the meat supply chain, which include selecting suppliers, placing orders, negotiating delivery options, invoicing, and monitoring logistics. 
With the exception of order placement, where users must complete the order form and initiate the process by clicking the 'Launch' button, all actions within this supply chain are executed automatically.
\begin{figure}[t]
    \centering
    \includegraphics[width=\textwidth]{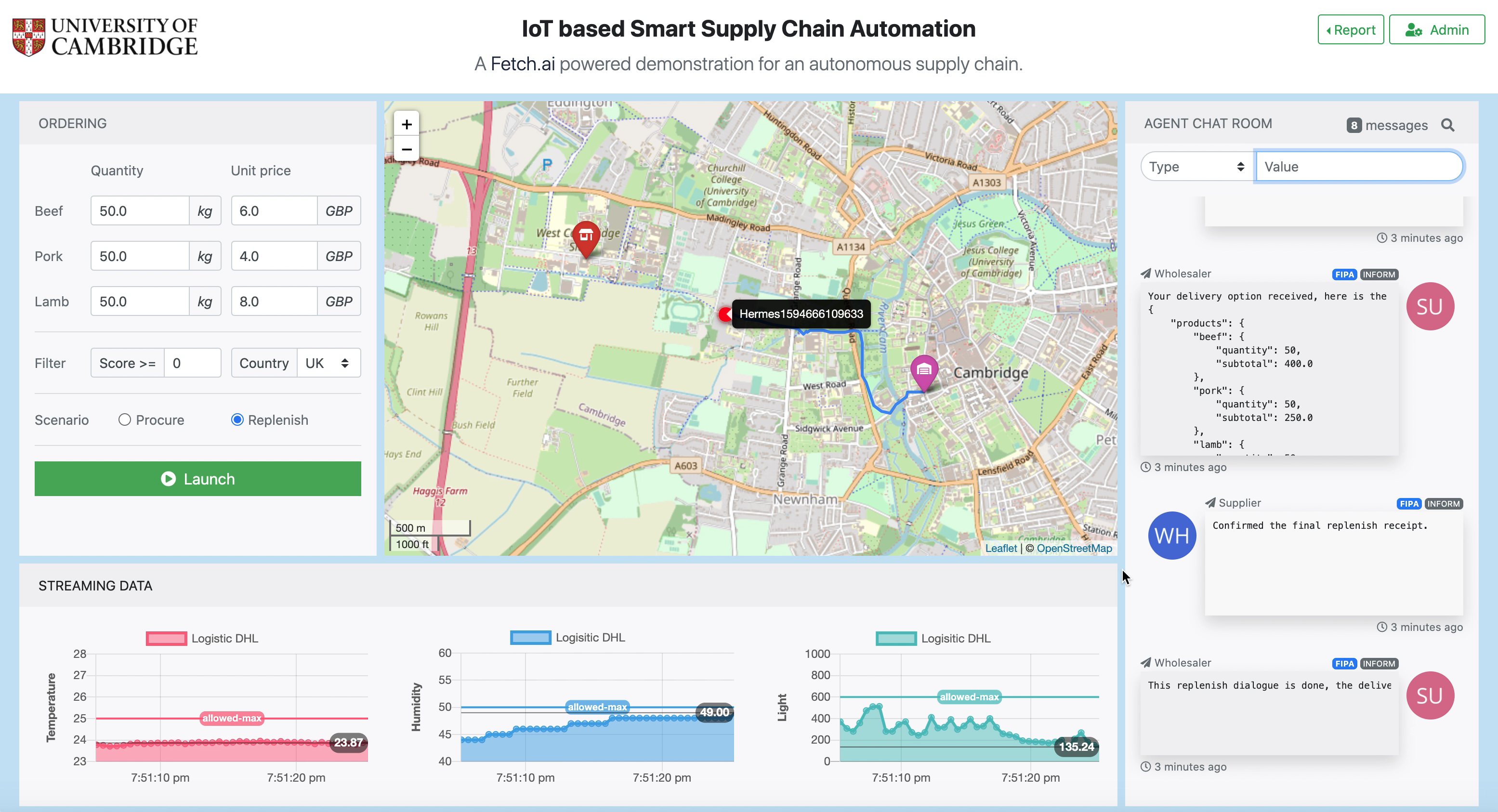}
    \caption{A screenshot of the system interface during a replenishment process. 
    In the central map area, a delivery vehicle is indicated by a red solid circle, positioned at around the midpoint of its journey along the route outlined in blue. 
    A infotip with black background is visible, displaying the delivery tracking number. 
    In the bottom area of the interface, real-time IoT readings for temperature, humidity, and illumination are shown, providing insights into the ambient conditions of the products being transported.
    In the right side of the interface, the chat history between all involved agents is displayed, visualising the processes of negotiation and coordination among stakeholders in the supply chain.}
    \label{fig:interface_middle}
\end{figure}

\subsection{Showcase}\label{sec:showcase}
We demonstrate the functionality of this ASC system by showcasing an automated meat procurement process.
\autoref{fig:interface_startup} shows the startup interface of the developed system.
As shown in \autoref{fig:interface_startup}, all four panels of this system are in their initial state with empty content and default values. 
For example, the system can order three types of meat: chicken, beef, and lamb, each with a default ordering quantity of 50kg. 
\autoref{fig:interface_middle}, on the other hand, presents the system's interface during its running, specifically, in the middle of a delivery during the CMC's replenishment process.

We showcase the ASC system using the default settings. 
After clicking the ``Launch'' button, the system starts the replenishment process autonomously, without requiring human intervention. 
Specifically, this process involves procuring specified quantities of meat products (50 kg each of pork, lamb, and beef) from selected suppliers and transporting them to the CMC using the logistics services provided by a designed 3PL provider. 
This process include various automated functions, such as supplier selection, inventory updating, logistics arrangement, transportation monitoring, and delivery service assessment.  
These functions are executed in a predefined order, triggered when prerequisite conditions are met. 
The automation involved in this process is facilitated by the backend multi-agent system, where autonomous agents act on behalf of the respective supply chain entities they represent. 
These agents including meat suppliers, the CMC, logistics companies, and 3PLs. 
These agents represent structural entities (described in \autoref{sec:definition}), gathering necessary data and making decisions.

This system also includes a wholesale process, in which the CMC supplies meat products to local retailers.
While there are variations in the selection of 3PLs, the wholesale process shares many similarities with the replenishment process. 
Therefore, we omit the description of this process.
When connected sequentially, these two automated processes represent a common goal of supply chain management, facilitating the movement of products from the source, through an intermediary (the CMC), to their ultimate destination.

Although these two processes are simulated and do not involve the physical movement of products, this system effectively illustrates the architectural guidelines for developing ASCs.
This system can be broadly separated into the five layers described in the MISSI model.
In the instrumentation layer, a set of three sensors were deployed to gather real-time environmental data, offering the upper layers insights into the surrounding conditions of the products.
Subsequently, these data were processed and transformed into standardised formats for various purposes, such as controlling product ambient conditions and assessing delivery services, which may involve data exchange. 
These data handling activities constitute the standardisation layer. 
Standardisation enables interconnection and interoperability.
This system adopted a MAS approach to provides a mechanism for connecting distributed and heterogeneous entities and objects. 
This approach enables connectivity and, consequently, facilitates data exchange and collaboration.
Integration is further facilitated by allowing the functions of different structure entities to achieve a coherent and streamlined process. 
In this system, this is achieved through interaction among agents and their internal units via the Contract Net protocol \citep{smith1981frameworks} --- a task-sharing protocol in multi-agent systems. 
Built upon these four layers, automation and intelligence-related technologies can be adopted to realise autonomous supply chain functions. 
These functions collectively form the manifestation layer.
In this system, functions within this layer, such as supplier selection, are more ``automated'' than ``intelligent'' since their decision-making relies on predefined rules. 
However, these functions are interconnected and coordinated to logically construct a loosely-coupled yet coherent system. 
By following the five layer conceptual design, the developed ASC system, albeit with limitations, demonstrate the feasibility of applying the proposed theories for the design and development of ASC systems.

\section{Discussion and Considerations}\label{sec:discussion}
The previous section presents a cast study, an ASC system implementation employing a MAS approach. 
This MAS approach provides a framework that logically links physically distributed entities within a supply chain, forming an integrated structure.
Within this structure, representative agents interact with others through messaging to ensure coherence.
It is important to note that this system falls considerably short of being considered as a complete autonomous system, let alone a {\it fully-fledged} realistic ASC system. 
Instead, it is an experimental proof-of-concept system based on the MIISI model.
Its implementation mainly focuses on three infrastructure layers of the model: instrumentation, interconnection, and integration, which are enabled by the adopted MAS approach.
The remaining two layers, standardisation and manifestation, receive comparatively less attention in this implementation.

This work is an initial attempt to create a {\it realistic} ASC system based on the proposed conceptual model.
According to the autonomy manifold presented in \autoref{fig:autonomy_manifold}, this system falls within the {\it automation-skewed} region.
However, both the intelligence and automation levels of the system are currently low.
The intelligence dimension mainly focuses on decision-making.
In current system, decision-making is based on a straightforward set of rules that consider only a limited set of factors. 
For example, one of these rules involves taking price into account when selecting potential suppliers. 
To effectively tackling real-world challenges in the supply chain domain, it is imperative to enhance the decision-making capabilities, enabling them to make informed decisions and propose appropriate courses of actions.
Regarding the automation dimension, the system mainly focuses on automating data flow and process execution.
However, this automation is constrained, only occurring along predefined data transmission paths. 
Additionally, as discussed in \autoref{sec:case_study}, the system does not consider financial and product flow, which are an integral part of a real-world supply chain.
A successful ASC system shall comprehensively tackle both of these two dimensions. 
This means enhancing both decision-making and automation capabilities, extending autonomy across a broad spectrum of supply chain activities, which may include automatic context-based decision making and the automated handling of physical flow through advanced robotics.

In addition to the technological aspects relevant to ASC development, advancing the ASC agenda needs addressing other crucial design considerations.
These include:

\begin{itemize}[nosep]
    \item {\it Cyber Security}:
    The adoption of technologies to enhance connectivity along the supply chain exposes companies to increased cyber risks.
    Organisations must adopt new and effective risk management strategies and tools to proactively prepare for potential cyber threats \citep{wef2017impact}.
    ASC systems, being highly interconnected, are particularly vulnerable to cyber attacks, which can trigger disruptions propagating throughout the entire supply chain.
    Consequently, ASC development requires proactive measures to safeguard against various cyber attacks. 
    
    \item {\it Data Security and Privacy}:
    The ASC era will witness the collection, access, and exchange of vast and diverse data of various types. 
    These data may include sensitive information critical to a company's core competitive advantage or restricted to authorised individuals and entities.
    As such, the design of ASC must incorporate robust mechanisms to ensure data security and privacy, protecting it from unauthorised access and malicious threats.
    
    \item {\it Trustworthiness}:
    ASCs entail collaboration among multiple entities responsible for managing the supply chain, each driven by its own objectives and potentially engaged in competition with others.
    Agents that represent these entities, including those utilising machine learning models such as BERT \citep{devlin2018bert}, GPT-3 \citep{raffel2020exploring}, and Llama \citep{touvron2023llama} for decision making, must be reliable and trustworthy.
    Their collaborative actions should result in mutual benefits for all supply chain participants, cultivating trust among them. 
    
    \item {\it Platform Neutrality}:
    ASC system development may rely on co-created or vendor platforms for delivering various services.
    These platforms must maintain neutrality, treating all parties impartially and refraining from favouring any specific entity
    This ensures fairness and transparency within the supply chain ecosystem, fostering healthy competition and collaboration among participants.
\end{itemize}

To address these challenges outlined above, an ASC system must implement a reliable and resilient architecture, along with effective interaction mechanisms.
Additionally, it should establish appropriate data access and authorisation policies.

In ASC development, managerial and cultural aspects are crucial considerations.
The transition to ASCs involves automating many manual processes, which has raised concerns about potential job losses across the supply chain \citep{blueyonder2020autonomous}.
However, the proliferation of connected and smart supply chain technologies is transforming employees into strategic decision-makers, as highlighted in a report by \citep{blueyonder2020autonomous}.
With routine tasks increasingly managed by autonomous solutions, employees can redirect their focus towards strategic activities or new roles.
This transformation results in a shift in the nature of required job roles. 
Companies should proactively design programmes to empower employees, providing them with the skills and expertise to work effectively with autonomous systems (or intelligent agents) and adapt to the evolving demands of their roles.

Moreover, a survey conducted by \citet{dhl2017supply} highlights the growing importance of leadership and strategic management skills in the future.
Given the multitude of autonomous functions and systems within ASCs, supply chain managers must prioritise these skills in strategic planning.
The ASC roadmap should also include small and middle-sized enterprises (SMEs), which play a significant role in the supply networks of large companies.
Unlike larger companies with substantial R\&D budgets, SMEs often have limited resources for investing in new technologies.
Therefore, it is crucial to develop appropriate technologies and offer governmental and policy support to ensure SMEs can participate in ASC development.
Additionally, legal and regulatory frameworks must be established to govern ASC-related activities, including intellectual property (IP), human-robot collaboration, and data security and privacy.

\section{Conclusion and Future Work}\label{sec:conclusion}
Modern supply chains have become increasingly networked, which not only heightens the cascading of risks but also complicates risk management \citep{butner2010smarter, christopher2016logistics}.
Moreover, the turbulent and uncertain environments in recent years exacerbated this situation. 
To adapt to disruptions, supply chains must evolve to be digitalised, more automated, smarter, and structurally flexible and resilient.
These requirements call for the new supply chain model: the autonomous supply chain (ASC), a self-governing supply chain with minimal or even no human intervention. 
Equipped with automation and intelligent decision-making capabilities, ASCs can self-adjust and promptly respond to uncertainties. 

As a concept that has been evolved over many years, systematic studies on the conceptual development of ASCs remain lacking, especially in contrast to the technical explorations that adopt various modern technologies (AI, IoT, robotics, etc) to digitally enhance supply chain tasks.
This paper thus aims to bridge this knowledge gap.

Specifically, we have provided a {\bf formal definition} of ASC, based on a newly defined concept of {\bf structural entities}. 
To better conceptualise ASC, we presented a two-dimensional {\bf autonomy manifold} to examine the autonomy development of a supply chain entity.
Building upon these characteristics, we proposed a conceptual model, {\bf the MIISI model}, which conceptualises ASCs as composed of five abstract and downward dependent layers: 
{\it Manifestation},
{\it Integration}, 
{\it Interconnection}, 
{\it Standardisation}, and {\it Instrumentation}.
Importantly, this model is agnostic to specific technologies, allowing for different implementations in each layer.
Additionally, we introduced the {\bf auxiliary concepts} of {\it internal connection} and {\it external connection}, with the latter categorised into three distinct types within the context of ASCs: {\it ad hoc}, {\it temporary}, and {\it established connections}.

Achieving a fully autonomous supply chain is an {\it incremental} journey that requires both technological and managerial efforts.
To support this journey, we introduced a seven-level supply chain autonomy reference model designed to assess a company's current supply chain autonomy level (SCAL).
These seven levels are a structured roadmap, providing companies with a reference trajectory for progressively achieving full supply chain autonomy.
In addition to these theoretical efforts, we briefly presented an ASC implementation case study on the meat supply chain.
This implementation, employing the MAS approach, showcases the realisation of an ASC system based on the proposed conceptual framework.

This study is among the first to formally address the conceptual development of ASCs, even as the term has gained attention for many years in news and industry reports. 
The conceptual artefacts presented in this paper, including the ASC definition, MIISI model, auxiliary concepts, and SCAL reference model, together provide a valuable foundation for future research and implementations on ASCs.
We anticipate that this study will inspire further research, both conceptual and technical, thereby contributing to the continual evolution of ASCs.
As the backbone of any business, supply chains with autonomy can certainly facilitate full information integration between multiple distributed supply chain entities. 
We thus expect this study to also contribute to the industrial information integration community by providing conceptual frameworks for integrating information through the lens of supply chains.

This work is an initial effort toward supply chain autonomy, and future work is needed to investigate this emerging area.
Our future work will be threefold: 
1) we will continue to interrogate the proposed conceptual constructs and conduct qualitative studies on their use in real-world SCM practices; 
2) we will extend our research by implementing ASCs at higher SCALs in new case studies across more industrial sectors; and 
3) we will investigate the potential of leveraging large language models to enhance the decision-making capabilities of agents within the context of ASCs.

\section*{acknowledgements}
This work was funded by the Research England's Connecting Capability Fund (grant number: CCf18-7157): Promoting the Internet of Things via Collaboration between HEIs and Industry (Pitch-In) and the EPSRC Connected Everything Network Plus under grant EP/S036113/1.

\bibliographystyle{abbrvnat}  
\bibliography{references}

\end{document}